\documentclass[11pt]{article}
\usepackage{coling2020}
\usepackage{times}
\usepackage{url}
\usepackage{latexsym}
\usepackage[textsize=tiny]{todonotes}
\usepackage{booktabs}
\usepackage{amsmath,amssymb,amsfonts,latexsym,dsfont,xspace}
\usepackage{physics,nicefrac}
\usepackage{multirow}
\usepackage{verbatim}
\usepackage{xcolor}
\usepackage{graphicx}
\usepackage{subcaption}
\usepackage{array}
\usepackage{pgfplotstable}
\usepackage{wrapfig}
\usepackage{cleveref} %

\makeatletter
\pgfplotstableset{
    discard if not/.style 2 args={
        row predicate/.code={
            \def\pgfplotstable@loc@TMPd{\pgfplotstablegetelem{##1}{#1}\of}
            \expandafter\pgfplotstable@loc@TMPd\pgfplotstablename
            \edef\tempa{\pgfplotsretval}
            \edef\tempb{#2}
            \ifx\tempa\tempb
            \else
                \pgfplotstableuserowfalse
            \fi
        }
    }
}
\makeatother

\colingfinalcopy %

\title{Is MAP Decoding All You Need?\\The Inadequacy of the Mode in Neural Machine Translation}

\author{Bryan Eikema \\
  University of Amsterdam \\
  {\tt b.eikema@uva.nl} \\\And
  Wilker Aziz \\
  University of Amsterdam \\
  {\tt w.aziz@uva.nl} \\}

\date{}

\newcommand{\E}[2]{\mathbb{E}_{#1}\left[#2\right]}
\newcommand{\thetamle}{\theta_{\text{MLE}}}

\DeclareMathOperator{\Cat}{Cat}

\DeclareMathOperator{\Exp}{Exp}
\DeclareMathOperator{\Poisson}{Poisson}
\DeclareMathOperator{\Gamm}{Gamma}
\DeclareMathOperator{\Dir}{Dir}
\DeclareMathOperator{\Multi}{Multinomial}

\DeclareMathOperator*{\argmax}{argmax} %

\newcommand{\flores}[0]{\textsc{Flores}\xspace}

\newcommand{\one}{\textit{i)}\xspace}
\newcommand{\two}{\textit{ii)}\xspace}
\newcommand{\three}{\textit{iii)}\xspace}

\newcommand{\eg}{\textit{e.g.}\xspace}
\newcommand{\ie}{\textit{i.e.}\xspace}
\newcommand{\wrt}{w.r.t.\xspace}

\begin{document}
\maketitle
\begin{abstract}
Recent studies have revealed a number of pathologies of neural machine translation (NMT) systems. Hypotheses explaining these mostly suggest there is something fundamentally wrong with NMT as a model or its training algorithm, maximum likelihood estimation (MLE). Most of this evidence was gathered using maximum \emph{a posteriori} (MAP) decoding, a decision rule aimed at identifying the highest-scoring translation, \ie the mode. We argue that the evidence corroborates the inadequacy of MAP decoding more than casts doubt on the model and its training algorithm. In this work, we show that translation distributions do reproduce various statistics of the data well, but that beam search strays from such statistics. We show that some of the known pathologies and biases of NMT are due to MAP decoding and not to NMT's statistical assumptions nor MLE. In particular, we show that the most likely translations under the model accumulate so little probability mass that the mode can be considered essentially arbitrary. We therefore advocate for the use of decision rules that take into account the translation distribution holistically. We show that an approximation to minimum Bayes risk decoding gives competitive results confirming that NMT models do capture important aspects of translation well in expectation.
\end{abstract}

\blfootnote{

    \hspace{-0.65cm}  %
    This work is licensed under a Creative Commons 
    Attribution 4.0 International Licence.
    Licence details:
    \url{http://creativecommons.org/licenses/by/4.0/}.

}

\section{Introduction} 
Recent findings in neural machine translation (NMT) suggest that modern translation systems have some serious flaws. %
This is based on observations such as: \one translations produced via beam search typically under-estimate sequence length \cite{sountsov-sarawagi-2016-length,koehn-knowles-2017-six}, the \emph{length bias};  \two translation quality generally deteriorates with better approximate search \cite{koehn-knowles-2017-six,murray-chiang-2018-correcting,ott-etal-2018-analyzing,kumar-sarawagi-2019-calibration}, the \emph{beam search curse}; 
\three the true most likely translation under the model (\ie, the mode of the distribution) is empty in many cases~\cite{stahlberg-byrne-2019-nmt} and a general negative correlation exists between likelihood and quality beyond a certain likelihood value~\cite{ott-etal-2018-analyzing}, we call this the \emph{inadequacy of the mode problem}.

A number of hypotheses have been formulated to explain these observations.
They mostly suggest there is something fundamentally wrong with NMT as a model 
(\ie, its factorisation as a product of locally normalised distributions) 
or its most popular training algorithm (\ie, regularised maximum likelihood estimation, MLE for short).
These explanations make an unspoken assumption, namely, that identifying the mode of the distribution, also referred to as maximum \emph{a posteriori} (MAP) decoding~\cite{lspbook}, is in some sense the obvious decision rule for predictions.
While this assumption makes intuitive sense and works well in unstructured classification problems,  it is less justified in NMT, where oftentimes the most likely translations
together account for very little probability mass, 
a claim we shall defend conceptually and provide evidence for in experiments. 
Unless the translation distribution is extremely peaked about the mode for every plausible input, criticising the model in terms of properties of its mode can at best say something about the adequacy of MAP decoding. %
Unfortunately, as previous research has pointed out, this is seldom the case \cite{ott-etal-2018-analyzing}.
Thus, pathologies about the mode  cannot be unambiguously ascribed to NMT as a model nor to MLE, and inadequacies about the mode cannot rule out the possibility that the model captures important aspects of translation well in expectation.

In this work, we criticise NMT models as probability distributions estimated via MLE in various settings: varying language pairs, amount of training data, and test domain. 
We observe that the induced probability distributions represent statistics of the data well in expectation, and that some length and lexical biases are introduced by approximate MAP decoding.
We demonstrate that beam search outputs are rare events, particularly so when test data stray from the training domain. %
The empty string, shown to often be the true mode  \cite{stahlberg-byrne-2019-nmt}, too is an infrequent event.
Finally, we show that samples obtained by following the model's own generative story are of reasonable quality, %
which suggests we should base decisions on statistics gathered from the distribution holistically. 
One such decision rule is minimum Bayes risk (MBR) decoding~\cite{goelbyrnembr,kumar-byrne-2004-minimum}.
We show that an approximation to MBR performs rather well, especially so when models are more uncertain.

To summarise: 
we argue that \one MAP decoding is not well-suited as a decision rule for MLE-trained NMT;
we also show that \two pathologies and biases observed in NMT are not necessarily inherent to NMT as a model or its training objective, rather, MAP decoding is at least partially responsible for many of these pathologies and biases;
finally, we demonstrate that \three a straight-forward approximation to a sampling-based decision rule known as minimum Bayes risk decoding gives good results, showing promise for research into decision rules that take into account the distribution holistically.

\section{Observed Pathologies in NMT} %
\label{sec:related}

Many studies have found that NMT suffers from a \emph{length bias}: NMT underestimates length which hurts the adequacy of translations. \newcite{cho-etal-2014-properties}
already demonstrate 
that NMT systematically degrades in performance for longer sequences. \newcite{sountsov-sarawagi-2016-length} identify the same bias in a chat suggestion task and argue that sequence to sequence models underestimate the margin between correct and incorrect sequences due to local normalisation. Later studies have also confirmed the existence of this bias in NMT~\cite{koehn-knowles-2017-six,stahlberg-byrne-2019-nmt,kumar-sarawagi-2019-calibration}.

Notably, all these studies employ beam search decoding. In fact, some studies link the length bias to the \emph{beam search curse}: the observation that large beam sizes hurt performance in NMT~\cite{koehn-knowles-2017-six}. \newcite{sountsov-sarawagi-2016-length} already note that larger beam sizes exacerbate the length bias. Later studies have confirmed this connection~\cite{blain2017exploring,murray-chiang-2018-correcting,yang-etal-2018-breaking,kumar-sarawagi-2019-calibration}. \newcite{murray-chiang-2018-correcting} attribute both problems to local normalisation which they claim introduces label bias~\cite{lafferty-etal-2001-conditional} to NMT. \newcite{yang-etal-2018-breaking} show that likelihood negatively correlates with translation length. 
These findings suggest that the mode might suffer from length bias, likely thereby failing to sufficiently account for adequacy. In fact, \newcite{stahlberg-byrne-2019-nmt}
show that 
oftentimes the true mode is the empty sequence.

The connection with the length bias is not the only reason for the beam search curse. \newcite{ott-etal-2018-analyzing} find that the presence of copies in the training data  cause the model to assign too much probability mass to copies of the input,
and that with larger beam sizes this copying behaviour becomes more frequent. %
\newcite{cohen-beck-2019-empirical} show that translations obtained with larger beam sizes often consist of an unlikely prefix with an almost deterministic suffix and are of lower quality.
In open-ended generation, \newcite{zhang-etal-2020-trading} correlate model likelihood with human judgements for a fixed sequence length, thus eliminating any possible length bias issues. 
They find that likelihood generally correlates positively with human judgements, up until an inflection point, after which the correlation becomes negative. 
An observation also made in translation with BLEU rather than human judgements~\cite{ott-etal-2018-analyzing}. We call this general failure of the mode to represent good translations in NMT the \emph{inadequacy of the mode problem}. 

\section{NMT and its Many Biases}
\label{sec:background}
MT systems are trained on sentence pairs drawn from a parallel corpus.
Each pair consists of a sequence $x$ in the source language and a sequence $y$ in the  target language. %
Most NMT models are conditional models \cite{cho-etal-2014-learning,bahdanau-etal-2015-nmt,transformer},\footnote{Though fully generative accounts do exist~\cite{shah-barber-2018,eikema-aziz-2019-auto}.} 
that is, only the target sentence is given random treatment. 
Target words are drawn in sequence from a product of locally normalised Categorical distributions
without Markov assumptions: $Y_j|\theta, x, y_{<j} \sim \Cat(f(x, y_{<j}; \theta))$.
At each step, a neural network $f(\cdot; \theta)$ maps from the source sequence $x$ and the prefix sequence $y_{<j}$ to the parameters of a Categorical distribution over the vocabulary of the target language.  %
These models are typically trained via regularised maximum likelihood estimation, MLE for short, where we search for the parameter $\thetamle$ that assigns maximum (regularised) likelihood to a dataset of observations $\mathcal{D}$. 
A local optimum of the MLE objective can be found by stochastic gradient-based optimisation  \cite{robbins-monro-1951-a,bottou-cun-2003-large}. %

For a trained model with parameters $\thetamle$ and a given input $x$, a translation is predicted by searching for the mode of the distribution: the sequence $y^\star$ that maximises $\log p(y|x, \thetamle)$. This is a decision rule also known as maximum \emph{a posteriori} (MAP) decoding~\cite{lspbook}.\footnote{The term MAP decoding was coined in the context of generative classifiers and their structured counterparts, where the posterior probability $p(y|x, \theta) \propto p(y|\theta)p(x|y, \theta)$ updates our prior beliefs about $y$ in light of $x$. This is not the case in NMT, where we do not express a prior over target sentences, and $p(y|x,\theta)$ is a direct parameterisation of the likelihood, rather than a posterior probability inferred via Bayes rule. Nonetheless, we stick to the conventions used in the MT literature.} 
Exact MAP decoding is intractable in NMT, and the beam search algorithm~\cite{sutskever-etal-2014-sequence} is employed as a viable approximation. %

It has been said that due to certain design decisions NMT suffers from a number of biases. %
We review those biases here %
and then discuss in Section~\ref{sec:mode-bias} one bias that has received very little attention and which, we argue, underlies many biases in NMT and explains some of the pathologies discussed in Section~\ref{sec:related}.

\paragraph{Exposure bias.} MLE parameters are estimated conditioned on observations sampled from the training data. 
Clearly, those are not available at test time, when we search through the learnt distribution.
This mismatch between training %
and test, 
known as exposure bias~\cite{ranzato-etal-2016-sequence}, has been linked to many of the  pathologies of NMT and  motivated modifications or alternatives to MLE aimed at exposing the model to its own predictions  during training~\cite{BengioEtAl2015,ranzato-etal-2016-sequence,shen-etal-2016-minimum,wiseman-rush-2016-sequence,zhang-etal-2019-bridging}. 
While exposure bias has been a point of critique mostly against MLE, it has only been studied in the context of approximate MAP decoding. The use of MAP decoding and its approximations shifts the distribution of the generated translations away from data statistics (something we provide evidence for in later sections), thereby exacerbating exposure bias.

\paragraph{Non-admissible heuristic search bias.} In beam search, partial translations are ranked in terms of log-likelihood without regards to (or with crude approximations of) their future score, which may lead to good translations being pruned too early. 
This corresponds to searching with a non-admissible heuristic \cite{AStar}, that is, a heuristic that may underestimate the likelihood of completing a translation.
This biased search affects statistics of beam search outputs in unknown ways and may well account for some
of the pathologies of Section~\ref{sec:related}, and has motivated variants of the algorithm aimed at comparing partial translations more fairly~\cite{huang-etal-2017-finish,shu-nakayama-2018-improving}.
This problem has also been studied in parsing literature, where it's known as imbalanced probability search bias \cite{ACL2020:CCG}.

\paragraph{Label bias.} Where a conditional model makes independence assumptions about its  inputs (\ie, variables the model does not generate), local normalisation prevents the model from revising its decisions, a problem known as \emph{label bias}  \cite{bottou-etal-1991-une,lafferty-etal-2001-conditional}.
This is a model specification problem which limits the  distributions a model can represent \cite{andor-etal-2016-globally}.
While this is the case in incremental parsing \cite{stern-etal-2017-effective} and simultaneous translation \cite{gu-etal-2017-learning}, where  inputs are incrementally  available for conditioning, 
this is \emph{not} the case in standard NMT~\cite[Section~5]{sountsov-sarawagi-2016-length}, where  inputs are available for conditioning in all generation steps.
It is plausible that local normalisation might affect
the kind of local optima we find in NMT, but that is orthogonal to label bias. %

\section{Biased Statistics and the Inadequacy of the Mode\label{sec:mode-bias}}

In most NMT research, criticisms of the model are based on observations about the mode, or an approximation to it obtained using beam search.
The mode, however, is not an unbiased summary of the probability distribution that the model learnt. 
That is, properties of the mode say little about properties of the learnt distribution (\eg, a short mode does not imply the model underestimates average sequence length).
MAP decoding algorithms and their approximations bias the statistics by which we criticise NMT.
They restrict our observations about the model to a single or a handful of outcomes which on their own can be rather rare. %
To gain insight about the model as a distribution, it seems more natural to use all of the information available to us, namely, all samples we can afford to collect, and search for frequent patterns in these samples. 
Evidence found that way better represents the model and its beliefs.

On top of that, the sample space of NMT is high-dimensional and highly structured. 
NMT models must distribute probability mass over a massive sample space (effectively unbounded). 
While most outcomes ought to be assigned negligible mass, for the total mass sums to $1$, the outcomes with non-negligible mass might still be too many.
The mode might only account for a tiny portion of the probability mass, and can actually be extremely unlikely under the learnt distribution.
Using the mode for predictions makes intuitive sense in unstructured problems, where probability distributions are likely very peaked, and in models trained with large margin methods \cite{vapnik_statistical_1998}, since those optimise a decision boundary directly.
With probability distributions  that are very spread out, and where the mode represents only a tiny bit of probability mass, targeting at the mode for predictions is much less obvious, an argument that we shall reinforce with experimental results throughout this analysis.\footnote{This perhaps non-intuitive notion that the most likely outcomes are rare and do not summarise a model's beliefs well enough is related to an information-theoretic concept, that of typicality \cite[Section 4.4]{mackay_information_2003}.}

At the core of our analysis is the concept of an unbiased sample from the model, which we obtain by ancestral sampling: iteratively sampling from distributions of the form $\Cat(f(x, y_{<j}; \theta))$, each time extending the generated prefix $y_{<j}$ with an unbiased draw, until the end-of-sequence symbol is generated.
By drawing from the model's probability distribution, unlike what happens in MAP decoding, we are imitating the model's training procedure.
Only we replace samples from the data by samples from the model, thus shedding light onto the model's fit.
That is, if these samples do not reproduce statistics of the data, we have an instance of poor fit.\footnote{Where one uses (approximate) MAP decoding instead of ancestral sampling this is known as exposure bias.} %
Crucially, ancestral sampling is not a pathfinding algorithm, %
thus the non-admissible heuristic search bias it not a concern. 
Ancestral sampling is \emph{not} a decision rule either, thus   returning a single sample as a prediction is not expected to outperform MAP decoding (or any other rule).
Samples can be used to diagnose model fit, as we do in Section~\ref{sec:fit}, and to approximate decision rules, as we do in Section~\ref{sec:mbr}.
In sum, we argue that MAP decoding is a source of various problems and that it biases conclusions about NMT. Next, we provide empirical evidence for these claims.

\section{Data \& System}
\label{sec:data-system}

We train our systems on German-English (de-en), Sinhala-English (si-en), and Nepali-English (ne-en), in both directions. 
For German-English we use all available WMT'18  \cite{bojar-etal-2018-findings} parallel data, except for Paracrawl, amounting to about $5.9$ million sentence pairs, 
and train a Transformer base model \cite{transformer}.
For Sinhala and Nepali, for which very 
little parallel data are available,  %
we  mimic the data and system setup of \newcite{guzman-etal-2019-flores}. 
As we found that the data contained many duplicate sentence pairs,
we removed duplicates, 
but left in those where only one side (source or target) of the data is duplicate to allow for paraphrases.
For all language pairs, we do keep a portion of the training data ($6,000$ sentence pairs) separate as held-out data for the analysis. 
In this process we also removed any sentence that corresponded exactly to either the source or target side of a held-out sentence from the training data. %
To analyse performance outside the training domain, we use WMT's \emph{newstest2018} for German-English, and the \flores datasets collected by \newcite{guzman-etal-2019-flores} for the low-resource pairs.
Our analysis is focused on MLE-trained NMT systems. However, as Transformers are commonly trained with label smoothing (LS) \cite{szegedy2016rethinking}, we do additionally report automatic quality assessments of beam search outputs on LS-trained systems.

\section{Assessing the Fit of MLE-Trained NMT}
\label{sec:fit}

\begin{figure}
    \centering
    \includegraphics[width=\linewidth]{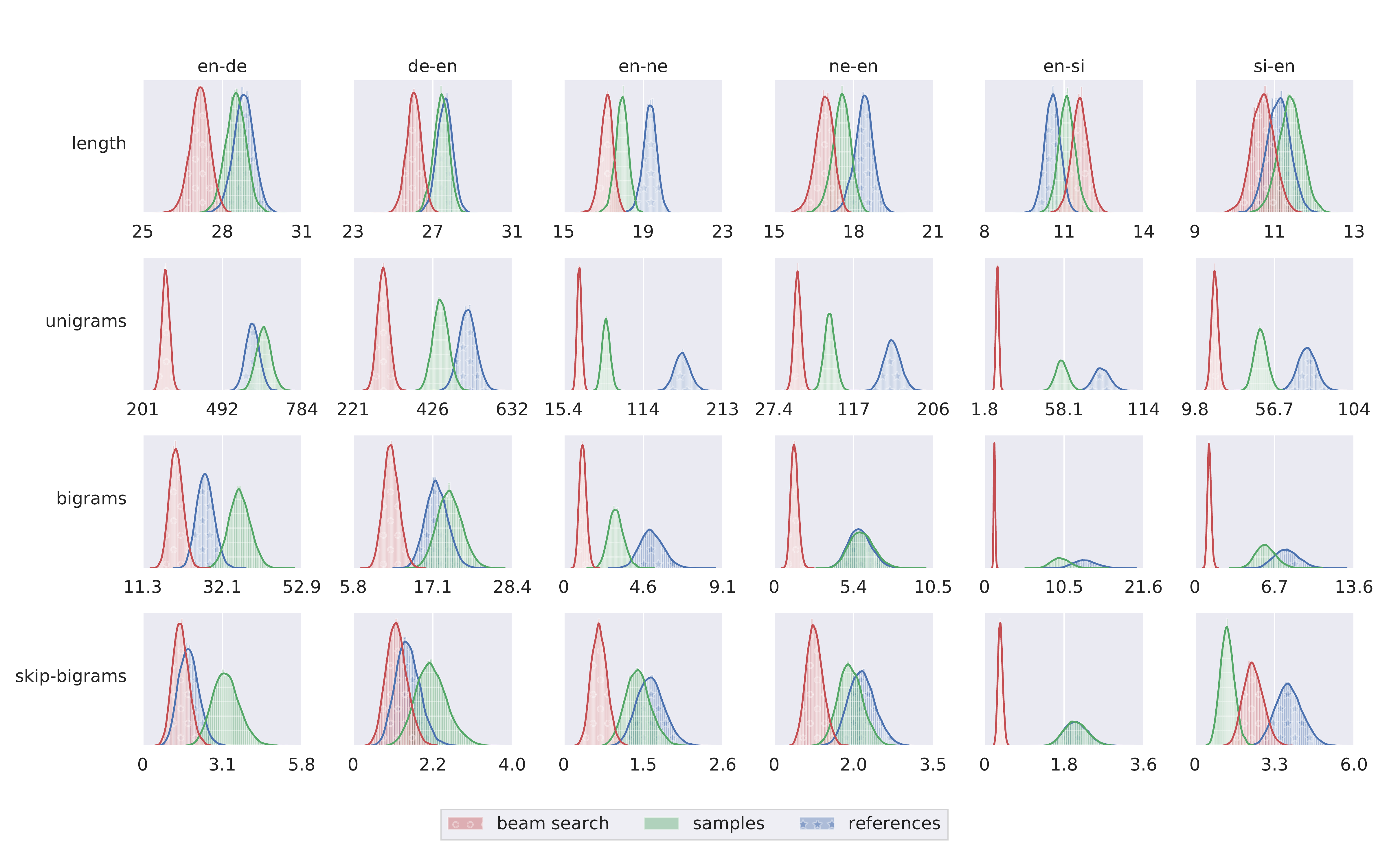}
    \caption{A comparison using hierarchical Bayesian models of statistics extracted from beam search outputs, samples from the model and gold-standard references.
    We show the posterior density on the y-axis, and the mean Poisson rate (length) and agreement with training data (unigrams, bigrams, skip-bigrams) on the x-axis for each group and language pair.
    }
    \label{fig:bda}
\end{figure}

We investigate the fit of the NMT models of Section~\ref{sec:data-system} on a held-out portion of the training data. This allows us to criticise MLE without confounders such as domain shift. We will turn to data in the test domain (\emph{newstest2018}, \flores) in Section~\ref{sec:set}. %
We compare unbiased samples from the model to gold-standard references and analyse statistics of several aspects of the data. If the MLE solution is good, we would expect statistics of sampled data to closely match statistics of observed data.

We obtain statistics from reference translations, ancestral samples, and beam search outputs and model them using hierarchical Bayesian models.
For each type of statistic, we formulate a joint model over these three groups and %
inspect the posterior distribution over the parameters of the analysis model. %
We also include statistics extracted from the training data in our analysis, and model the three \emph{test groups} as a function of posterior inferences based on training data statistics.
Our methodology follows that advocated by \newcite{bda3} and \newcite{blei-2014-build}.
In particular, 
we formulate separate hierarchical models to inspect length, lexical, and word order statistics: sequence length, unigram and bigram counts, and skip-bigram counts, respectively.\footnote{Skip-bigrams are pairs of tokens drawn in the same order as they occur in a sentence, but without enforcing adjacency.}
In Appendix~\ref{sec:app:analysis-models}, we describe in detail all analysis models, inference procedures, and predictive checks that confirm their fit. 

For length statistics, we look at the expected posterior Poisson rate for each group, each rate can be interpreted as that group's average sequence length. Ideally, the expected Poisson rates of predicted translations are close to those of gold-standard references. 
Figure~\ref{fig:bda} (top row) shows the inferred posterior distributions for all language pairs. 
We observe that samples generated by NMT capture length statistics reasonably well, overlapping a fair amount with the reference group. In almost all cases we observe that beam search outputs stray away from data statistics, usually resulting in shorter translations.

For unigrams, bigrams, and skip-bigrams, we compare the posterior agreement with training data of each group (this is formalised in terms of a scalar concentration parameter whose posterior we can plot). Higher values indicate a closer resemblance to training data statistics. For each statistic, the posterior distribution for gold-standard references gives an indication of ideal values of this agreement variable. 
Figure~\ref{fig:bda} (rows 2--4) show all posterior distributions. 
In most cases the gold-standard references agree most with the training data, as expected, followed by samples from the model, followed by beam search outputs. For nearly all statistics and language pairs beam search outputs show least agreement with the training data, even when samples from the model show similar agreement as references do. Whereas samples from the model do sometimes show less similarity than references, in most cases they are similar and thus lexical and word order statistics are captured reasonably well by the NMT model. Beam search on the other hand again strays from training data statistics, compared to samples from the model.

\section{Examining the Translation Distribution}
\label{sec:set}
The NMT models of Section~\ref{sec:data-system} specify complex distributions over an unbounded space of translations. Here we examine properties of these distributions by inspecting translations in a large set of unbiased samples. 
To gain further insight we also analyse our models in the test domain (\emph{newstest2018}, \flores).

\subsection{Number of Likely Translations}
\label{sec:size}

\begin{figure}
    \centering
    
  \centering
  \includegraphics[width=\linewidth]{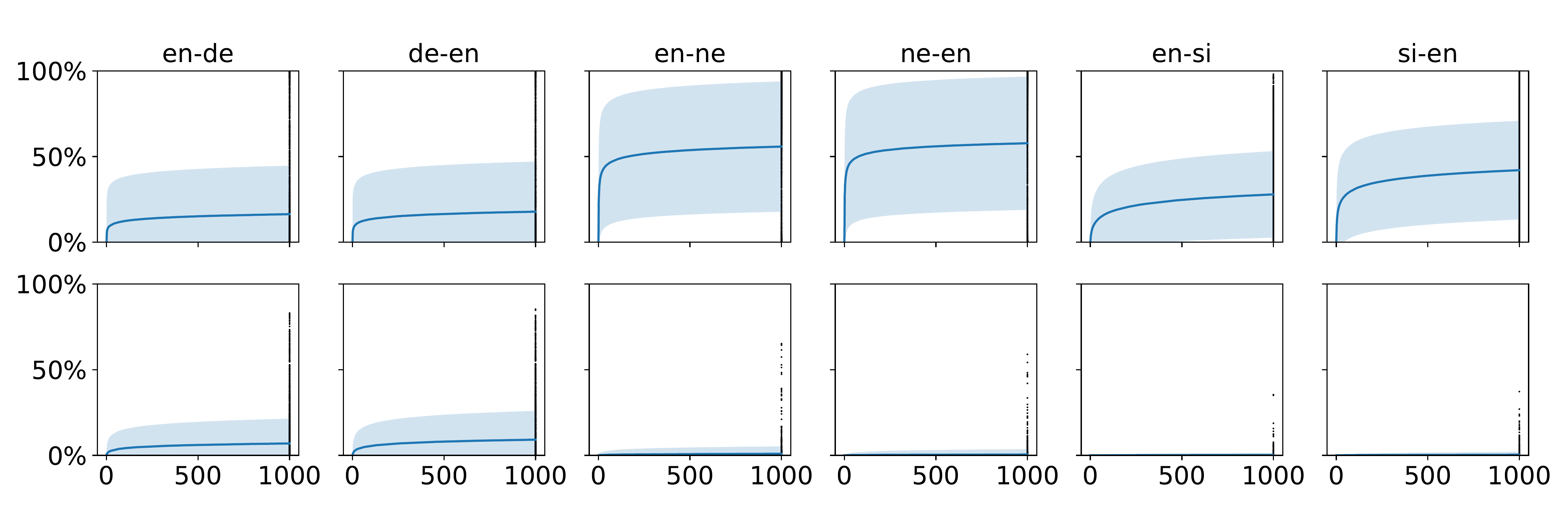}

    \caption{Cumulative probability of the unique translations in 1,000 ancestral samples on the held-out (top), and \emph{newstest2018} / \flores (bottom) test sets. The dark blue line shows the average cumulative probability over all test sentences, the shaded area represents 1 standard deviation away from the average. The black dots to the right show the final cumulative probability for each individual test sentence.}
    \label{fig:size}
\end{figure}

NMT, by the nature of its model specification, assigns probability mass to each and every possible sequence consisting of tokens in its vocabulary. Ideally, however, a well-trained NMT model assigns the bulk of its probability mass to good translations of the input sequence. %
We take $1,000$ unbiased samples from the model for each input sequence and count the cumulative probability mass of the unique translations sampled. Figure~\ref{fig:size} shows the average cumulative probability mass for all test sentences with $1$ standard deviation around it, as well as the final cumulative probability values for each input sequence. %
For the held-out data we observe that, on average, between  $16.4\%$ and $57.8\%$ of the probability mass is covered. %
The large variance around the mean shows that in all language pairs we can find test sentences for which nearly all or barely any probability mass has been covered after $1,000$ samples.
That is, even after taking $1,000$ samples, only about half of the probability space has been explored. 
The situation is much more extreme when translating data
from the test domain (see bottom half of Figure~\ref{fig:size}).\footnote{For English-German and German-English the test domain would not be considered out-of-domain here, as both training and test data concern newswire.}  %
Naturally, the NMT model is much more uncertain in this scenario, and this is very clear from the amount of probability mass that has been covered by $1,000$ samples: on average, only between $0.2\%$ and $0.9\%$ for the low-resource pairs and between $6.9\%$ and $9.1\%$ for English-German of the probability space has been explored. This shows that the set of likely translations under the model is very large and the probability distribution over those sentences mostly quite flat, especially so in the test domain.
In fact, if we look at each input sequence individually, we see that for $37.0\%$ (en-de), $35.5\%$ (de-en), $18.5\%$ (en-ne), $15.7\%$ (ne-en), $9.2\%$ (en-si), and $3.3\%$ (si-en)  of them all $1,000$ samples are unique. On the test domain data these numbers increase to $46.7\%$ (en-de), $41.5\%$ (de-en), $52.1\%$ (en-ne), $86.8\%$ (ne-en), $84.6\%$ (en-si), and $87.3\%$ (si-en). For these input sequences, the translation distributions learnt are so flat that in these $1,000$ samples no single translation stands out over the others.

\subsection{Sampling the Mode}
\label{sec:sampling-mode}

As the predominant decision rule in NMT is MAP decoding, which we approximate via beam search, it is natural to ask how frequently it is that the beam search output is observed amongst unbiased samples.
We find that the beam search output is contained within $1,000$ unbiased samples for between $54.3\%$ and $92.2\%$ of input sequences on the held-out data. In the test domain, we find that on English-German for between $44.3\%$ and $49.3\%$, and in the low-resource setting for between $4.8\%$ and $8.4\%$ of the input sequences the beam search output is contained in the set. This shows that for a large portion of the input sequences, the beam search solution 
is thus quite a rare outcome under the model.

Recently, \newcite{stahlberg-byrne-2019-nmt} showed that oftentimes the true mode of a trained NMT system is the empty sequence. This is worrying since NMT decoding is based on mode-seeking search. We find that for between $7.2\%$ and $29.1\%$ of input sequences for held-out data and between $2.8\%$ and $33.3\%$ of input sequences in the test domain an empty sequence is sampled at least once in $1,000$ samples. When an empty sequence is sampled it only occurs on average $1.2 \pm 0.5$ times. %
Even though it could well be, as the evidence that \newcite{stahlberg-byrne-2019-nmt} provide is strong, that often the true mode under our models is the empty sequence, the empty string remains a rather unlikely outcome under the models.

\subsection{Sample Quality}
\label{sec:automatic-eval}

\begin{table}[]
    \centering
    \pgfplotstableread{data/automatic-eval-coling-horizontal.txt}\loadedtable
\pgfplotstablesort[
    sort key={order}, 
    sort cmp={string <},
]{\sortedtable}{\loadedtable}
\pgfplotstabletypeset[
    columns={%
        lp, %
        ls_ID,
        beam_ID,sample_ID,mbr_m_ID,oracle_m_ID,
        ls_OD,beam_OD,sample_OD,mbr_m_OD,oracle_m_OD
    },
    columns/lp/.style={
        column name={Task},
        string type,
        column type=l,
    },
    columns/ls_ID/.style={
        column name=LS,
        column type=r,
        precision=1,
        clear infinite,
        fixed,
        fixed zerofill,
    },
    columns/beam_ID/.style={
        column name=beam,
        column type=r,
        precision=1,
        clear infinite,
        fixed,
        fixed zerofill,
    },
    columns/sample_ID/.style={
        column name=sample,
        column type=r,
        precision=1,
        clear infinite,
        fixed,
        fixed zerofill,
    },    
    columns/mbr_m_ID/.style={
        column name={MBR},
        column type=r,
        precision=1,
        clear infinite,
        fixed,
        fixed zerofill,
    },
    columns/oracle_m_ID/.style={
        column name={Oracle},
        column type=r,
        precision=1,
        clear infinite,
        fixed,
        fixed zerofill,
    },
   columns/ls_OD/.style={
        column name=LS,
        column type=r,
        precision=1,
        clear infinite,
        fixed,
        fixed zerofill,
    },
    columns/beam_OD/.style={
        column name=beam,
        column type=r,
        precision=1,
        clear infinite,
        fixed,
        fixed zerofill,
    },
    columns/sample_OD/.style={
        column name=sample,
        column type=r,
        precision=1,
        clear infinite,
        fixed,
        fixed zerofill,
    },    
    columns/mbr_m_OD/.style={
        column name={MBR},
        column type=r,
        precision=1,
        clear infinite,
        fixed,
        fixed zerofill,
    },
    columns/oracle_m_OD/.style={
        column name={Oracle},
        column type=r,
        precision=1,
        clear infinite,
        fixed,
        fixed zerofill,
    },
    every head row/.style={ 
        before row={
            \toprule 
            & \multicolumn{5}{c}{Training Domain} 
            & \multicolumn{5}{c}{Test Domain} \\
            \cmidrule(r{1pt}){2-6} 
            \cmidrule(l{1pt}){7-11} 
        },
        after row=\midrule,
    },
    every last row/.style={
        after row={
            \midrule 
            \multicolumn{1}{l}{High-resource }
            & 23.1 & 23.1 & 18.6 & 22.7 & 26.0 & 37.4 & 37.1 & 23.6 & 34.4 & 38.3 
            \\
            \multicolumn{1}{l}{Low-resource}
            & 32.7 & 32.6 & 29.5 & 34.0 & 39.1 & 25.9 & 24.3 & 21.8 & 26.0 & 28.9
            \\
            \multicolumn{1}{l}{All}
            & 29.5 & 29.4 & 25.9 & 30.2 & 34.8 & 29.7 & 28.6 & 22.4 & 28.8 & 32.0
            \\ \bottomrule
        }
    },
]\sortedtable

    \caption{METEOR scores under different strategies for prediction: beam search,  single sample, MBR, and an oracle rule. MBR and the oracle both use $30$ ancestral samples and sentence-level METEOR as utility, but the oracle has access to the reference.
    To show that our MLE-trained systems are competitive with LS-trained systems, we list the LS column (using beam search). The sample columns show average scores of 30 independent samples from the model. All standard deviations were below 0.2.
    }
    \label{tab:automatic-eval}
\end{table}

\begin{figure}[t]
    \centering
    
  \includegraphics[width=\linewidth]{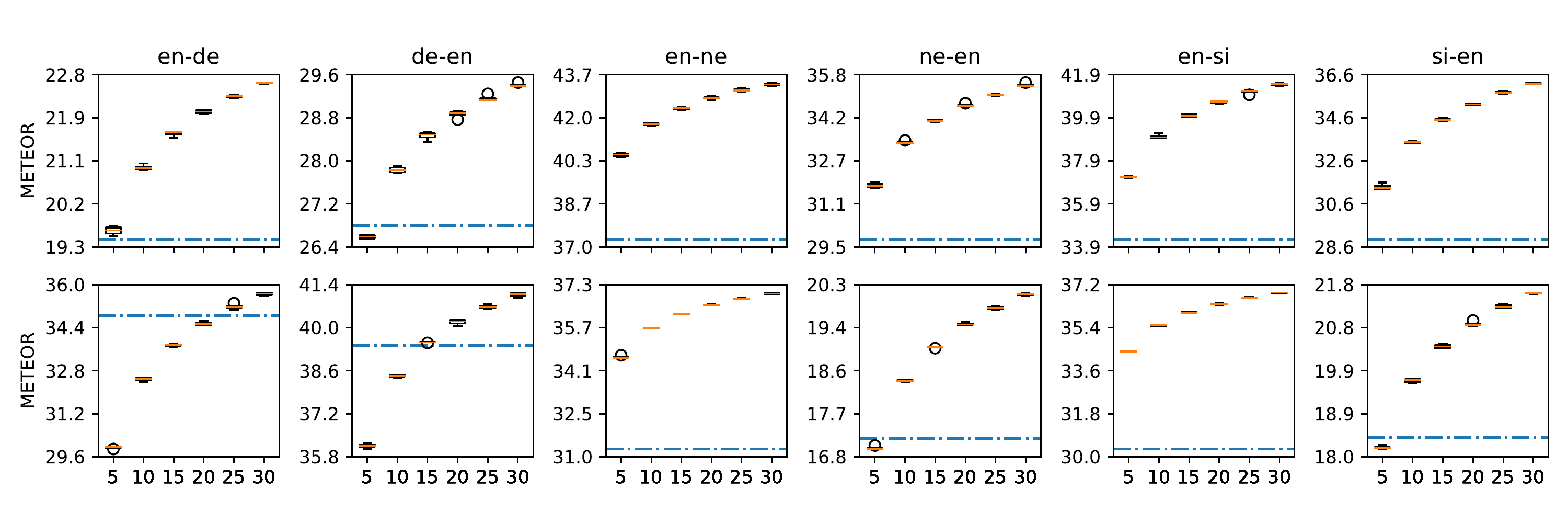}

    \caption{METEOR scores for oracle-selected samples as a function of sample size on the held-out data (top) and \emph{newstest2018} / \flores (bottom) test sets. For each sample size we repeat the experiment $4$ times and show a box plot per sample size. Dashed blue lines show beam search scores.}
    \label{fig:oracle}
\end{figure}

The number of translations that an NMT model assigns non-negligible mass to can be very large as we have seen in Section~\ref{sec:size}. We now investigate what the average quality of these samples is. For quality assessments, we compute 
 METEOR~\cite{meteor} using the \texttt{mteval-v13a} tokeniser.\footnote{For our analysis, it is convenient to use a metric defined both at the corpus and at the segment level. We use METEOR, rather than BLEU \cite{bleu}, for it outperforms (smoothed) BLEU at the segment-level \cite{ma-etal-2018-results}.}
We translate the test sets using a single ancestral sample per input sentence and repeat the experiment $30$ times to report the average in 
Table~\ref{tab:automatic-eval} (sample). We also report beam search scores (beam). %
We see that, on average, samples of the model always perform worse than beam search translations. This is no surprise, of course, as ancestral sampling is not a fully fledged decision rule, but simply a technique to unbiasedly explore the learnt distribution. Moreover, beam search itself does come with some adjustments to perform well (such as a specific beam size and length penalty). The gap between sampling and beam search is between %
$0$ and $14.4$ METEOR. 
The gap can thus be quite large, but overall the quality of an average sample is reasonable compared to beam search. We also observe that the variance of the sample scores is small with standard deviations below $0.2$.
Next, we investigate the performance we would achieve if we could select the best sample from a set. %
For that, we employ an oracle selection procedure using sentence-level METEOR with the reference translation to select the best sample from a set of samples. We vary sample size from $5$ to $30$ samples and repeat each experiment four times. Figure~\ref{fig:oracle} plots the results in terms of corpus-level METEOR. Average METEOR scores for oracle selection out of $30$ samples are shown in Table~\ref{tab:automatic-eval}.
METEOR scores steadily increase with sample size. For a given sample size we observe that variance is generally very small. %
Only between $5$ and $10$ samples are required to outperform beam search in low-resource language pairs and English-German in the training domain, but surprisingly $15$ to $25$ samples are necessary for English-German in the test domain.
Still, this experiment shows that samples are of reasonable and consistent quality with respect to METEOR. For fewer than $30$ random samples the model could meet or outperform beam search performance in most cases, if we knew how to choose the best sample from the set.
This is a motivating result for looking into sampling-based decision rules.

\subsection{Minimum Bayes Risk Decoding}
\label{sec:mbr}
We have seen that translation distributions %
spread mass over a large set of likely candidates, oftentimes without any clear preference for particular translations within the set (Section~\ref{sec:size}). 
Yet, this set is not arbitrary, it captures various statistics of the data well (Section~\ref{sec:fit}) and holds potentially good translations (Section~\ref{sec:automatic-eval}). 
Even though the model does not single out one clear winner, the translations it does assign non-negligible mass to share statistics that correlate with the reference translation.
This motivates a decision rule that exploits all information we have available about the distribution. In this section we explore one such decision rule: minimum Bayes risk (MBR) decoding.

For a given \emph{utility function} $u(y, h)$, which assesses a hypothesis $h$ against a reference $y$,  statistical decision theory~\cite{mbr} prescribes that the optimum decision $y^\star$ is the one that maximises expected utility (or minimises expected loss) under the model: $y^\star = \argmax_{h \in \mathcal H(x)} ~\mathbb E_{p(y|x,\theta)}[u(y, h)]$, 
where the maximisation is over the entire set of possible translations $\mathcal H(x)$.
Note that there is no need for a human-annotated reference, expected utility is computed by having the model \emph{fill in} reference translations.
This decision rule, known as MBR decoding in the NLP literature~\cite{goelbyrnembr}, is especially suited where we trust a model in expectation but not its mode in particular~\cite[Section 5.3]{lspbook}.\footnote{MAP decoding is in fact MBR with a very strict utility function which evaluates to $1$ if a translation exactly matches the reference, and $0$ otherwise.  As a community, we acknowledge by means of our evaluation strategies (manual or automatic) that exact matching is inadequate for translation, unlike many unstructured classification problems, admits multiple solutions.}
MBR decoding, much like MAP decoding, is intractable. 
We can at best obtain unbiased estimates of expected utility via Monte Carlo (MC) sampling,  %
and we certainly cannot search over the entirety of $\mathcal H(x)$. 
Still, a tractable approximation can be designed, albeit without any optimality guarantees.
We use MC both to approximate the support $\mathcal H(x)$ of the distribution and to estimate the expected utility of a given hypothesis.
In particular, we maximise over the support $\bar{\mathcal{H}}(x)$ of the empirical distribution obtained by ancestral sampling:
\begin{equation}
    y^\star = \argmax_{h \in  \bar{\mathcal{H}}(x)} ~ \frac{1}{S} \sum_{s=1}^S u(y^{(s)}, h) \quad\text{for } y^{(s)} \sim p(y|x,\theta) ~,
\end{equation}
which runs in time $\mathcal O(S^2)$.
Though approximate, this rule has interesting properties: MC improves with sample size, occasional pathologies in the set pose no threat, and there is no need for incremental search.

Note that whereas our translation distribution might be very flat over a vast number of translations, not showing a clear ordering in terms of relative frequency within a large set of samples, this need not be the case under expected utility. For example, in Section~\ref{sec:sampling-mode} we found that for some input sequences the empty sequence is contained within the 1,000 samples in our set and appears in there roughly once on average. If all the 1,000 samples are unique (as we found to often be the case in Section~\ref{sec:size}), we cannot distinguish the empty sequence from the other 999 samples in terms of relative frequency.
However, under most utilities the empty sequence is so unlike the other sampled translations that it would score very low in terms of expected utility.

We chose METEOR as utility function for it, unlike BLEU, is well-defined at the sentence level.\footnote{Even though one can alter BLEU such that it is defined at the sentence level (for example, by adding a small positive constant to $n$-gram counts), this ``smoothing'' in effect biases BLEU's sufficient statistics. Unbiased statistics are the key to MBR, thus we opt for a metric that is already defined at the sentence level.} %
We estimate expected utility using $S=30$ ancestral samples, and use the translations we sample to make up an approximation to $\mathcal H(x)$.
Results are shown in Table~\ref{tab:automatic-eval}.
As expected, MBR considerably outperforms the average single sample performance by a large margin and in many cases is on par with beam search, consistently outperforming it in low-resource pairs. 
For English-German in the test domain, we may need more samples to close the gap with beam search.
Whereas an out-of-the-box solution based on MBR requires further investigation, this experiment shows promising results. Crucially, it shows that exploring the model as a probability distribution holds great potential.

\section{Related Work}

Some of our observations have been made in previous work.  \newcite{ott-etal-2018-analyzing} observe that unigram statistics of beam search stray from those of the data, while random samples do a better job at reproducing them. 
\newcite{holtzman-etal-2019-the} find that beam search outputs have disproportionately high token probabilities compared to natural language under a sequence to sequence model.
Our analysis is more extensive, we include richer statistics about the data, more language pairs, and vary the amount of training resources, leading to new insights about MLE-trained NMT and the merits of mode-seeking predictions.

\newcite{ott-etal-2018-analyzing} also observe that NMT learns flat distributions, they analyse a high-resource English-French system trained on $35.5$ million sentence pairs from WMT'14 and find that after drawing %
$10,000$ samples from the WMT'14 validation set less than $25\%$ of the probability space has been explored.
Our analysis shows that even though NMT distributions do not reveal clear winners, they do emphasise translations that share statistics with the reference, which motivates us to look into MBR.

MBR decoding is old news in machine translation \cite{kumar-byrne-2004-minimum,tromble-etal-2008-lattice}, but it has received little attention in NMT. Previous approximations to MBR in NMT employ beam search to define the support and to evaluate expected utility (with probabilities renormalised to sum to $1$ in the beam), these studies report the need for very large beams \cite{stahlberg-etal-2017-neural,blain2017exploring,shu-nakayama-2017-later}.
They claim the inability to directly score better translations higher is a deficiency of the model scoring function. 
We argue this is another piece of evidence for the inadequacy of the mode: by using beam search, they emphasise statistics of high-scoring translations,  potentially rare and inadequate ones.
Very recently, \newcite{borgeaud-emerson-2020-leveraging} present a voting-theory perspective on decoding for image captioning and machine translation. Their proposal is closely-related to MBR, but motivated differently. Their decision rule too is guided by beam search, which may emphasise pathologies of highest-likelihood paths, but they also propose and investigate stronger utility functions which lead to improvements \wrt length, diversity, and human judgements.

The only instance that we are aware of where unbiased samples from an NMT model support a decision rule is the concurrent work by \newcite{naskar2020energybased}. The authors make the same observation that we make in Section~\ref{sec:automatic-eval}, namely that an oracle selection from a small set of samples of an NMT model shows great potential, greatly outperforming beam search. Motivated by this observation, the authors propose a re-ranking model that learns to rank sampled translations according to their oracle BLEU. Using the trained model to re-rank a set of 100 samples from the NMT model they find strong improvements over beam search of up to 3 BLEU points, again showing the potential of sampling-based decision rules.

\section{Conclusion}
\label{sec:discussion}

In this work, we discuss the inadequacy of the mode in NMT and question the appropriateness of MAP decoding. We show that for such a high dimensional problem as NMT, the probability distributions obtained with MLE are spread out over many translations, and that the mode often does not represent any significant amount of probability mass under the learnt distribution. We therefore argue that MAP decoding is not suitable as a decision rule for NMT systems. Whereas beam search performs well in practice, it suffers from biases of its own (\ie, non-admissible heuristic search bias), it shifts statistics away from those of the data  (\ie, exposure bias and other lexical and length biases), and in the limit of perfect search it falls victim to the inadequacy of the mode. 
Instead, we advocate for research into decision rules that take into account the probability distribution more holistically. 
Using ancestral sampling we can explore the learnt distribution in an unbiased way and devise sampling-based decision rules. Ancestral sampling does not suffer from non-admissibility, and, if the model fit is good, there is no distribution shift
either.

We further argue that criticisms about properties of the mode of an NMT system are not representative of the probability distributions obtained from MLE training.
While this form of criticism is perfectly reasonable where approximations to MAP decoding
are the only viable alternative, there are scenarios where we ought to criticise models as probability distributions.
For example, where we seek to correlate an observed pathology with a design decision, such as factorisation, or training algorithm. In fact, we argue that many of the observed pathologies and biases of NMT are at least partially due to the use of (approximate) MAP decoding, rather than inherent to the model or its training objective. %

Even though NMT models spread mass over many translations, we find samples to be of decent quality and contain translations that outperform beam search outputs even for small sample sizes, further motivating the use of sampling-based decision rules.
We show that an approximation to one such decision rule,
MBR decoding, shows competitive results. This confirms that while the set of likely translations under the model is large, the translations in it share many statistics that correlate well with the reference.

MLE-trained NMT models admit probabilistic interpretation and an advantage of the probabilistic framework is that a lot of methodology is already in place when it comes to model criticism as well as making predictions. We therefore advocate for criticising NMT models as probability distributions and making predictions using decision rules that take into account the distributions holistically. We hope that our work paves the way for research into scalable sampling-based decision rules and motivates researchers to assess model improvements to MLE-trained NMT systems from a probabilistic perspective.

\section*{Acknowledgements}

\begin{wrapfigure}[3]{l}{0.10\linewidth}
\vspace{-13pt}
\includegraphics[width=0.12\textwidth]{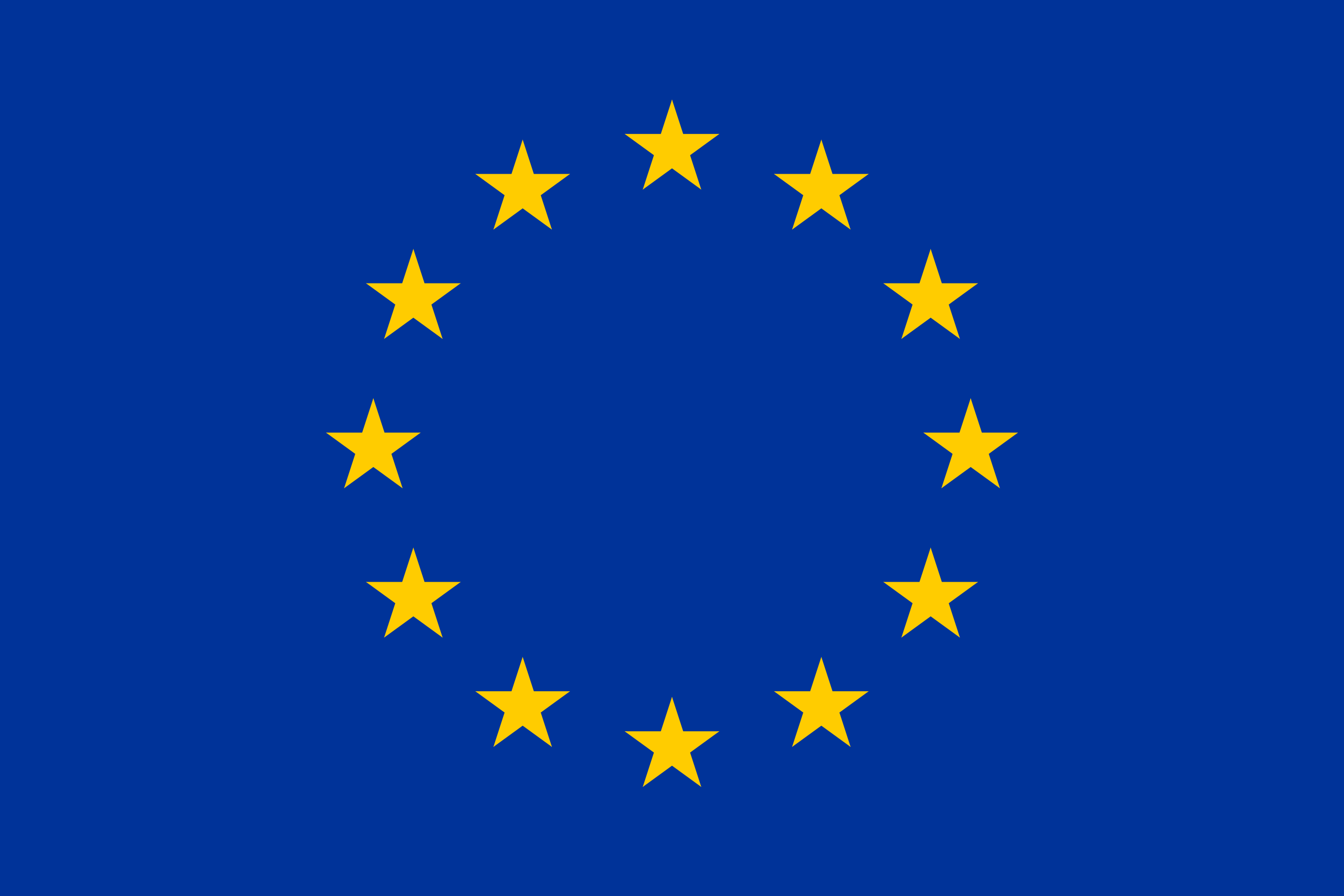}
\end{wrapfigure}

This project has received funding from the European Union's Horizon 2020 research and innovation programme under grant agreement No 825299 (GoURMET). We also thank Khalil Sima'an, Lina Murady, Milo\v{s} Stanojevi\'c, and Lena Voita for comments and helpful discussions.
A Titan Xp card used for this research was donated by the NVIDIA Corporation. 

\bibliographystyle{coling}
\bibliography{anthology,analysis}

\appendix

\section{Analysis Models}
\label{sec:app:analysis-models}

\subsection{Length Analysis}
\label{sec:app:length-model}
We model length data  from the training group using a hierarchical Gamma-Poisson model. Each target sequence length is modelled as being a draw from a Poisson distribution with a Poisson rate parameter specific to that sequence.
All Poisson rates share a common population-level Gamma prior with population-level parameters $\alpha$ and $\beta$. The population-level parameters are given fixed Exponential priors set to allow for a wide but reasonable range of Poisson rates \emph{a priori}.

\begin{align*}
    & \alpha \sim \Exp(1) & \beta \sim \Exp(10)\\
    & \lambda_i \sim \Gamm(\alpha, \beta) &  y_i \sim \Poisson(\lambda_i)
\end{align*}

Here, $i$ indexes one particular data point. This model is very flexible, because we allow the model to assign each datapoint its own Poisson rate. We model test groups as an extension of the training group. Test group data points are also modelled as draws from a Gamma-Poisson model, but parameterised slightly differently. 

\begin{align*}
    & \mu = \E{}{\Gamm(\alpha, \beta|\mathcal{D}_T)} & \eta \sim \Exp(1.)\\
    & s_g \sim \Exp(\eta) & t_g = 1/\mu \\
    & \lambda_{gi} \sim \Gamm(s_g, t_g) & y_{gi} \sim \Poisson(\lambda_{gi})
\end{align*}

Here, $i$ again indexes a particular data point, $g$ a group in $\{\text{reference}, \text{sampling}, \text{beam}\}$, and $\mathcal{D}_T$ denotes the data of the training group. All Poisson rates are individual to each datapoint in each group. The Poisson rates do share a group-level Gamma prior, whose parameters are $s_g$ and $t_g$.  $s_g$ shares a prior among all test groups and therefore ties all test groups together. $t_g$ is derived from posterior inferences on the training data by taking the expected posterior Poisson rate in the training data and inverting it. This is done such that the mean Poisson rate for each test group is $s_g \cdot \mu$, where $s_g$ can be seen as a parameter that scales the expected posterior training rate for each test group individually. 
We infer Gamma posterior approximations for all unknowns using stochastic variational inference (SVI). After inferring posteriors, we compare predictive samples to the observed data in terms of first to fourth order moments to verify that the model fits the observations well.

\subsection{Lexical \& Word Order Analyses}
\label{sec:app:lexical-and-word-order-model}
We model unigram and (skip-)bigram data from the training group using a hierarchical Dirichlet-Multinomial model as shown below:

\begin{align*}
    & \alpha \sim \Gamm(1, 1) & \beta \sim \Gamm(1, 1)\\
    & \theta \sim \Dir(\alpha) &  \psi_u \sim \Dir(\beta)\\
    & u \sim \Multi(\theta) & b|u \sim \Multi(\psi_u)\\
\end{align*}

Here, we have one Gamma-Dirichlet-Multinomial model to model unigram counts $u$, and a separate Dirichlet-Multinomial model for each $u$ (the first word of a bigram) that $b$ (the second word of a bigram) conditions on, sharing a common Gamma prior that ties all bigram models. This means that we effectively have $V+1$ Dirichlet-Multinomial models (where $V$ is BPE vocabulary size) in total to model the training group, where the $V$ bigram models share a common prior.

We model the three test groups using the inferred posterior distributions on the data of the training group $\mathcal{D}_T$. We compute the expected posterior concentration of the Dirichlets in the training group models and normalise it such that it sums to 1 over the entire vocabulary. The normalisation has the effect of spreading the unigram and bigram distributions. The test groups are modelled by scaling this normalised concentration parameter using a scalar. In order for test-groups to recover the training distribution the scaling variable needs to be large to undo the normalisation. This scalar, $s_g$ for unigrams or $m_g$ for bigrams, can be interpreted as the amount of agreement of each test group with the training group. The higher this scalar is, the more peaked the test group Multinomials will be about the training group lexical distribution.

\begin{align*}
    & \mu(\alpha) = \E{}{\alpha | \mathcal{D}_T}. %
    & \mu(\beta) = \E{}{\beta | \mathcal{D}_T}\\
    & \eta_s \sim \Gamm(1, 0.2) & \eta_m \sim \Gamm(1, 0.2)\\
    & s_g \sim \Gamm(1, \eta_s) & m_g \sim \Gamm(1, \eta_m)\\
    & \theta_g \sim \Dir(s_g \cdot \mu(\alpha)) &  \psi_g \sim \Dir(m_g \cdot \mu(\beta))\\
    & u_g \sim \Multi(\theta_g) & b_g|u_g \sim \Multi(\psi_g)\\
    & g \in \{\text{reference}, \text{sampling}, \text{beam}\}
\end{align*}

We do collapsed inference for each Dirichlet-Multinomial (as we are not interested in assessing $\theta_g$ or $\phi_g$), and infer posteriors approximately using SVI with Gamma approximate posterior distributions. To confirm the fit of the analysis model, we compare posterior predictive samples to the observed data in terms of absolute frequency errors of unigrams and bigrams as well as ranking correlation.

\end{document}